\PassOptionsToPackage{table}{xcolor}
\documentclass[]{fairmeta}
\usepackage{amsthm,amsmath,amssymb}
\usepackage{graphicx}
\usepackage[numbers,sort&compress]{natbib}
\usepackage{booktabs}
\usepackage{bm}
\usepackage{xspace}
\usepackage{tabularx} 
\usepackage{overpic}
\usepackage{pgf}

\usepackage{geometry} 
\usepackage{multirow}
\usepackage{enumitem}
\usepackage{fontspec}
\usepackage{sectsty}
\usepackage{caption}
\usepackage{subcaption}
\usepackage[ruled,linesnumbered]{algorithm2e}
\usepackage{amsmath}
\usepackage{amssymb}
\usepackage[hidelinks,breaklinks,colorlinks=true, linkcolor=blue, citecolor=red, urlcolor=cyan]{hyperref}
\usepackage{pifont}
\newcommand{\cmark}{\ding{51}}
\newcommand{\xmark}{\ding{55}}
\usepackage{booktabs}
\usepackage{caption}
\usepackage{subcaption}
\usepackage{graphicx}
\newfontfamily\calmregular[
    BoldFont={Outfit-Regular.ttf},
    BoldFeatures={FakeBold=1.4}
]{Outfit-Regular.ttf}
\newcommand{\calmfont}[1]{{{\calmregular #1}}}

\newtheorem{theorem}{Theorem}[]

\newtheorem{remark1}[theorem]{Remark}

\newcommand{\modelname}{HoloAgent-0\xspace}
\newcommand{\holonavi}{HoloNavi\xspace}
\newcommand{\holoagentnav}{HoloAgent-Nav\xspace}
\newcommand{\holomotion}{HoloMotion\xspace}
\newcommand{\holobrain}{HoloBrain\xspace}
\definecolor{green1}{RGB}{120, 198, 121} 
\definecolor{green2}{RGB}{173, 221, 142} 
\definecolor{green3}{RGB}{217, 240, 199} 
\definecolor{yellow1}{RGB}{255, 237, 160} 
\definecolor{yellow2}{RGB}{254, 217, 118} 
\definecolor{orange1}{RGB}{253, 174, 107} 
\definecolor{orange2}{RGB}{253, 141, 90}  
\definecolor{red1}{RGB}{252, 108, 96}     
\definecolor{red2}{RGB}{227, 74, 74}  
\newcommand{\getsem}[3]{%
    \pgfmathsetmacro{\normalized}{(#1-#2)/(#3-#2)}%
    \ifdim \normalized pt > 0.89pt \cellcolor{green1}%
    \else\ifdim \normalized pt > 0.78pt \cellcolor{green2}%
    \else\ifdim \normalized pt > 0.67pt \cellcolor{green3}%
    \else\ifdim \normalized pt > 0.56pt \cellcolor{yellow1}%
    \else\ifdim \normalized pt > 0.45pt \cellcolor{yellow2}%
    \else\ifdim \normalized pt > 0.34pt \cellcolor{orange1}%
    \else\ifdim \normalized pt > 0.23pt \cellcolor{orange2}%
    \else\ifdim \normalized pt > 0.12pt \cellcolor{red1}%
    \else \cellcolor{red2}\fi\fi\fi\fi\fi\fi\fi\fi%
}

\allsectionsfont{\calmfont}

\title{\calmfont{\modelname: A Unified Embodied Agent Framework with 3D Spatial Memory}}

\author[*,1]{\calmfont{Xiaolin Zhou}}
\author[*,1]{\calmfont{Liu Liu}}
\author[*,1]{\calmfont{Tingyang Xiao}}
\author[*,1]{\calmfont{Wei Feng}}
\author[2]{\calmfont{Fa Fu}}
\author[2]{\calmfont{Xinrui Meng}}
\author[1]{\calmfont{Xinjie Wang}}
\author[1]{\calmfont{Jialiang Han}}
\author[1]{\calmfont{Boyang Yu}}
\author[1]{\calmfont{Yun Du}}
\author[2]{\calmfont{Wei Sui}}
\author[1]{\calmfont{Zhizhong Su}}

\affiliation[1]{\calmfont{Horizon Robotics}}
\affiliation[2]{\calmfont{D-Robotics Robotics}}
\contribution[*]{Equal contribution.}

\code{\url{https://github.com/HorizonRobotics/HoloAgent}}
\webpage{\url{horizonrobotics.github.io/robot_lab/holoagent}}
\correspondence{Liu Liu at \email{nemo.liu@horizon.auto}}

\begin{document}

\abstract{
    LLM agents follow a practical execution loop in digital environments: they reason over structured states, invoke tools, inspect feedback, and revise actions. 
    Extending this loop to physical robots is difficult because physical execution is continuous, embodiment-dependent, uncertain, and constrained by safety. 
    Existing embodied-AI systems have advanced manipulation, spatial understanding, navigation, and humanoid control, but these capabilities often remain specialized modules or loosely coupled decision loops. 
    In this work, we introduce \modelname, a unified embodied agent framework for real-world robot deployment. 
    \emph{Embodied AgentOS} converts language instructions into executable skill graphs, schedules robot resources, monitors execution, and triggers clarification or re-planning from runtime feedback. 
    \modelname organizes heterogeneous robot models and controllers through three coupled layers: Embodied AgentOS for closed-loop execution, 3D spatial memory for physical-world grounding, and embodied skills for robot action.
    We deploy \modelname on real hardware and evaluate its spatial memory, long-horizon navigation, and closed-loop execution across motion generation, object search, cross-robot coordination, and mobile manipulation.
}

\maketitle
\vspace{-2mm}
\begin{figure}[ht]
\centering
\includegraphics[width=0.95\linewidth]{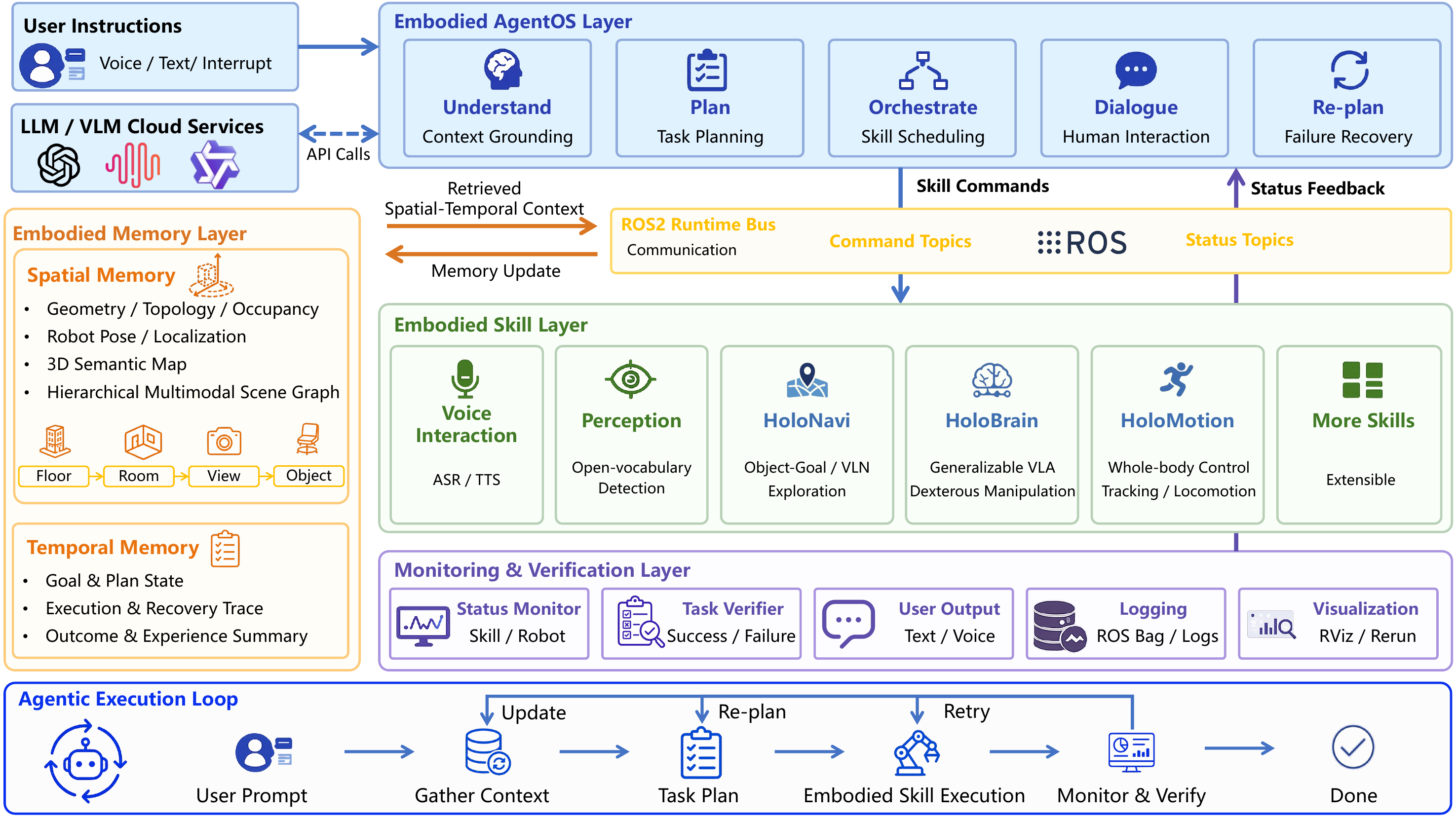}
\caption{\textbf{Overview of the \modelname framework.}
\modelname connects Embodied AgentOS, embodied memory, and embodied skills through a ROS2 command/status bus, forming a closed loop for spatial retrieval, skill-graph planning, execution monitoring, memory updates, and feedback-driven
re-planning.}
\label{fig:framework}
\end{figure}
\vspace{-4mm}

\section{Introduction}
\label{sec:intro}

In digital environments, Large Language Model (LLM) agents follow a practical execution loop: they reason over structured states, invoke software tools, inspect execution feedback, and revise subsequent actions~\citep{react,toolformer,reflexion,autogen,voyager}.
Transferring this loop to physical robots is substantially harder because physical execution is continuous, embodiment-dependent, uncertain, and constrained by safety~\citep{kaelbling2011hierarchical,garrett2021integrated,saycan,inner_monologue}.
This transfer exposes an \emph{embodiment gap}: physical skills do not behave like software APIs with clean input/output types, deterministic outputs, complete feedback, or reversible side effects.
A robot may navigate to the wrong room because spatial memory is stale, fail to grasp an object that has moved, or receive only partial progress from a locomotion controller.
The central challenge is therefore system-level: embodied agents need an execution abstraction that makes physical skills composable, observable, and verifiable during long-horizon task execution.

Recent embodied-AI systems have advanced many capability modules for physical agents, including generalizable manipulation, spatial understanding, navigation, and humanoid motion control.
Vision-language-action (VLA), video-action (VA), and whole-body action models provide general-purpose manipulation and interaction skills~\citep{pi0,bjorck2025gr00t,kim2024openvla,robovlm,spatialVLA,holobrain0,gr1,lingbotva,bi2025motus,dreamzero}; spatial understanding~\citep{scanqa,3dllm,embodiedscan,spatialvlm,vlm3r,spa3r,lin2025bip3d} and VLN models~\citep{zhang2024vision,mapdream,progress_think,monodream,aux_think} support navigation in complex environments; and motion-control models~\citep{gmt,any2track,sonic,holomotion1} enable diverse embodied movements.
Recent embodied multimodal systems further explore high-level coordination. For example, $\pi_{0.5}$ uses a dual-level inference procedure that predicts a textual subtask before generating low-level continuous actions~\citep{pi05,palm_e,rt2,bjorck2025gr00t}.
Nevertheless, many systems still treat manipulation, navigation, spatial understanding, and motion control as specialized modules, model-level policies, or loosely coupled agent loops.
They lack a physical execution interface that composes heterogeneous skills, grounds execution in persistent 3D memory, tracks partial progress, and triggers recovery from embodied failures.

We present \modelname, a unified embodied agent framework for real-world robot deployment.
Rather than replacing low-level policy models, \modelname organizes heterogeneous robot capabilities into a closed-loop workflow centered on \emph{Embodied AgentOS}.
AgentOS converts natural-language instructions into executable skill graphs, allocates robot resources, monitors execution, and re-plans from runtime feedback.

The framework consists of three coupled layers.
Embodied AgentOS provides task planning, scheduling, monitoring, and failure recovery; the Memory Layer maintains persistent spatial grounding and temporal execution history; and the Skill Layer exposes robot capabilities as typed actions with structured inputs, progress signals, failure modes, and recoverability status.
Through this interface, AgentOS can compose specialized backends such as \holonavi for navigation, \holomotion~\citep{holomotion1} for whole-body motion, and \holobrain~\citep{holobrain0} for manipulation.
Together, these layers connect cloud-level reasoning, on-device execution, persistent memory, and feedback-driven recovery in a unified embodied agent loop.

\begin{itemize}[leftmargin=*]

\item \textbf{\calmfont{An AgentOS runtime for physical robot execution.}}
We formulate long-horizon embodied task execution as a system-organization problem and introduce \emph{Embodied AgentOS}, a closed-loop workflow that converts language instructions into executable skill graphs, schedules robot resources, tracks task state, and re-plans from runtime feedback.

\item \textbf{\calmfont{A physical execution interface with spatial and temporal memory.}}
We organize robot capabilities through typed embodied skills, spatial and temporal memory, and runtime verification. This interface makes heterogeneous skills composable, grounds planning in persistent 3D spatial memory, preserves task history across execution, and supports monitoring and failure recovery.

\item \textbf{\calmfont{Real-world deployment and closed-loop evaluation.}}
We instantiate \modelname on real robot hardware, quantitatively evaluate its 3D semantic mapping and long-horizon navigation performance, and qualitatively demonstrate closed-loop motion control, object search, cross-robot coordination, and mobile manipulation across real deployments.

\end{itemize}

\section{Embodied AgentOS: A Closed-Loop Runtime for Robot Execution}
\label{sec:framework}

\begin{figure}[ht]
    \centering
    \includegraphics[width=\linewidth]{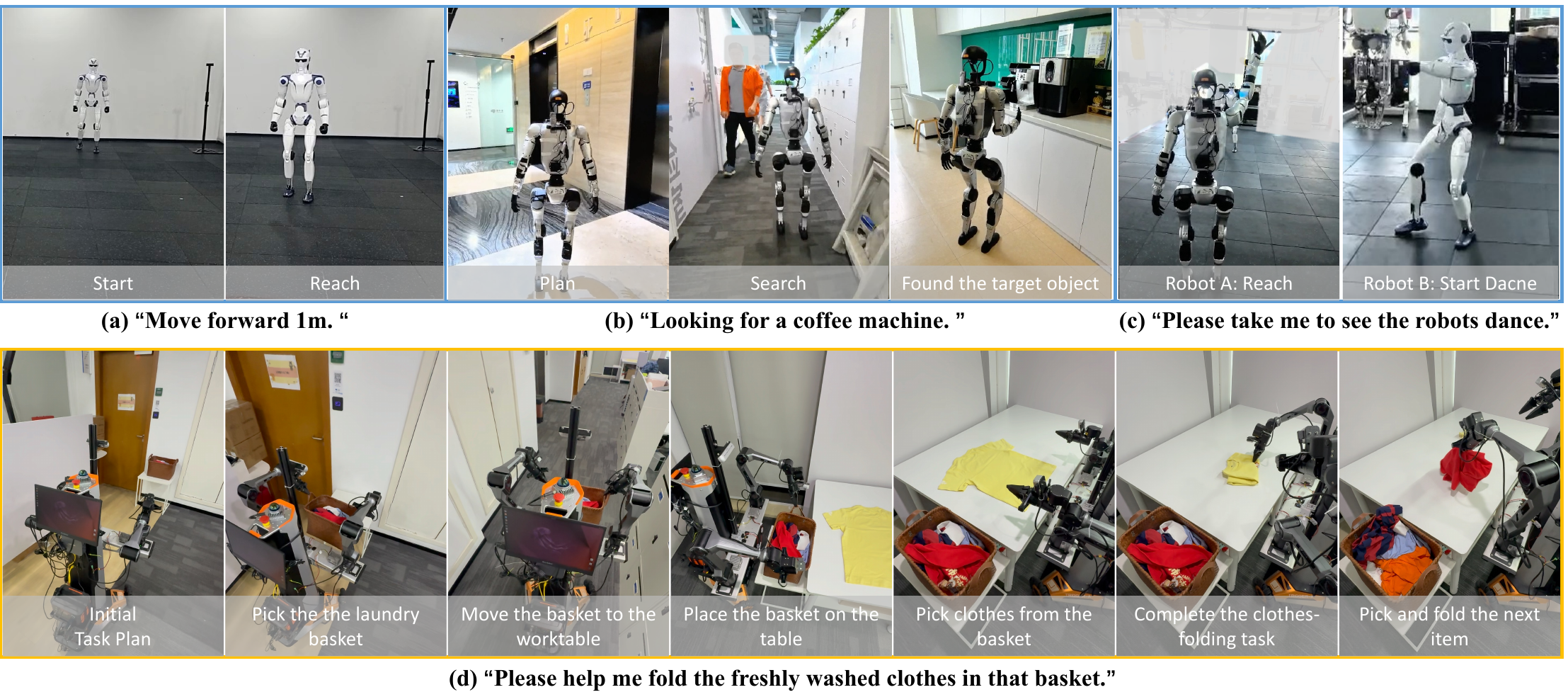}
    \caption{\textbf{Representative closed-loop executions with \modelname.}
    (a) \textbf{Prompt Motion Control}: execute and verify short-horizon whole-body commands.
    (b) \textbf{Active Object Search}: explore, build the map, and verify the target coffee machine.
    (c) \textbf{Cross-Robot Coordination}: route one robot while another performs a dance skill.
    (d) \textbf{Long-Horizon Mobile Manipulation}: decompose laundry folding into navigation, pick-and-place, motion, and manipulation steps.}
    \label{fig:framework_vis}
\end{figure}

Embodied AgentOS is the runtime layer in \modelname that turns natural-language intent into closed-loop robot execution. Rather than treating the language model as a one-shot planner, AgentOS maintains task state, retrieves spatial context from memory, schedules skill calls, monitors execution status, and revises the plan when observed outcomes diverge from the intended state. As shown in Fig.~\ref{fig:framework}, the runtime connects AgentOS, the Memory Layer, the Skill Layer, and monitoring/verification through a ROS2 command/status bus.
Figure~\ref{fig:framework_vis} illustrates representative behaviors enabled by this runtime.

\subsection{Design Principles}
\label{subsec:design_principles}

\modelname follows four design principles that adapt the language-agent loop to reliable robot execution:
\begin{itemize}[leftmargin=*]
    \item \textbf{\calmfont{Closed-loop first.}} AgentOS treats planning as a repeated observe--retrieve--act--monitor cycle rather than a one-shot text generation problem. This loop lets the system revise its plan when a skill fails, a goal is ambiguous, or the environment changes during execution.
    \item \textbf{\calmfont{Memory-centric.}} Persistent spatial and temporal memory provide the primary context for planning. Instead of relying only on the current camera view or dialogue context, AgentOS queries rooms, views, objects, poses, active goals, and recent skill outcomes from memory before dispatching actions.
    \item \textbf{\calmfont{Typed skill interface.}} Robot capabilities are exposed as typed, monitorable skill calls with command parameters and status events. This interface separates high-level task reasoning from embodiment-specific controllers while preserving the progress and failure evidence needed for recovery.
    \item \textbf{\calmfont{Observable by default.}} The runtime records and surfaces command/status events, retrieved memory, state transitions, and skill outcomes through user feedback, logs, and visualization. This observability supports real-robot debugging and gives AgentOS the evidence needed to verify or revise execution.
\end{itemize}



\subsection{Runtime Layers and ROS2 Execution Interface}
\label{subsec:operating_stack}

AgentOS implements the runtime loop through four functional layers connected by a ROS2 command/status interface. Command topics carry scheduled skill calls from AgentOS to robot capability modules, while status topics return progress, failures, sensor health, retrieved context, and memory-update events. This topic-level interface keeps the system modular: developers can replace individual models or controllers while preserving the closed-loop feedback path needed for long-horizon execution.

\textbf{\calmfont{AgentOS Layer.}}
The AgentOS layer turns each user instruction into an executable, monitorable skill graph. Given a voice or text command, AgentOS parses the request, retrieves task history and spatial evidence from memory, and decomposes the instruction into skill nodes with ordering, preconditions, and recovery dependencies. During execution, the scheduler dispatches ready skill calls, consumes status feedback, and triggers clarification or re-planning when the current plan becomes invalid.

\textbf{\calmfont{Skill Layer.}}
The embodied skill layer realizes the agent's body through callable robot capabilities, including interaction, perception, navigation, manipulation, and whole-body motion. Each skill consumes structured command parameters and publishes status events such as progress, success, failure mode, confidence, latency, and recoverability. The skill interface lets AgentOS compose heterogeneous robot capabilities without depending on the internal architecture of each backend.

\textbf{\calmfont{Memory Layer.}}
The memory layer provides persistent spatial and temporal memory shared by planning, localization, perception, and cross-embodiment collaboration. Spatial memory returns candidate objects, navigable regions, scene-graph context, and localization hypotheses. Temporal memory records active goal and plan state, execution and recovery traces, and outcome summaries. After execution, the memory-update module writes new observations and skill outcomes back into the corresponding spatial and temporal records.

\textbf{\calmfont{Monitoring \& Verification Layer.}}
The monitoring and verification layer converts raw execution evidence into decisions and feedback. During deployment, it verifies skill outcomes, exposes user-facing text or voice feedback, and provides signals for AgentOS re-planning. During development, it records ROS2 bags, structured logs, RViz views, and Rerun visualizations for diagnosing failures across perception, memory update, planning, and robot control.

Together, these layers define AgentOS as a closed-loop runtime rather than a single planner. The next section details the typed skill interface that this runtime dispatches and monitors.

\section{Skill Layer: Embodied Skills as the Executable Action Interface}
\label{sec:embodied_skill_layer}

The embodied skill layer forms the executable boundary between AgentOS plans and robot hardware. Rather than asking the language planner to emit low-level controls, \modelname exposes robot capabilities as typed, monitored skill calls. A skill call specifies what the robot should achieve, the embodiment-specific backend decides how to execute it, and the runtime status stream reports whether the intended state was reached. This interface gives AgentOS a stable action space while allowing foundation models, classical controllers, and hybrid systems to evolve independently behind the same ROS2 command/status interface.

\subsection{Typed Skill Calls and Runtime Status}
\label{subsec:skill_execution_interface}

We define a lightweight execution interface that turns robot capabilities into typed, monitorable skill calls. Unlike software tools, physical skills can fail partially, depend on embodiment constraints, and return delayed or incomplete feedback. The interface therefore specifies not only the command sent to a backend, but also the expected effect and the runtime evidence used by AgentOS to verify, retry, or re-plan.

\textbf{\calmfont{Command schema.}} Each skill declares a command name, typed parameters, preconditions, target references, and expected effects, e.g., \texttt{move\_to(room=kitchen)}, \texttt{pick(object=mug)}, or \texttt{speak(text=...)}. AgentOS uses these fields to construct skill graphs, bind memory-grounded targets, and check whether the observed outcome satisfies the intended postcondition.

\textbf{\calmfont{Runtime status interface.}} Each backend publishes an execution trace rather than a hidden success flag. The trace reports progress, success, failure mode, confidence, latency, and recoverability, allowing AgentOS to distinguish completed skills from blocked, ambiguous, unsafe, or recoverable executions. These status events become planning evidence for continuing, retrying, asking for clarification, updating memory, or re-planning.

\textbf{\calmfont{Embodiment-specific backend.}} A learned policy, a classical controller, or a hybrid system grounds the same skill interface on a specific body. This separation keeps the high-level action space stable across humanoids, mobile bases, and manipulators while allowing each platform to implement navigation, manipulation, interaction, or whole-body motion with its own sensing, actuation, and safety constraints.

The typed skill abstraction keeps the AgentOS loop independent of any specific embodiment or model checkpoint. AgentOS schedules symbolic skill calls, and the embodied skill layer grounds those calls in the concrete skill families described below.

\subsection{Interaction and Perception Skills}
\label{subsec:interaction_speech}

\textbf{\calmfont{Interaction skills.}}
As the user-facing control channel, the speech interface wraps ASR, dialogue state, and TTS into callable skills such as \texttt{listen} and \texttt{speak}. These skills return status events for transcription confidence, clarification requests, user confirmation, interruptions, and delivery completion, allowing AgentOS to keep the human in the loop when task intent or execution state is uncertain.

\textbf{\calmfont{Perception skills.}}
As the grounding channel between robot observations and 3D spatial memory, open-vocabulary perception exposes skills such as \texttt{detect}, \texttt{localize}, and \texttt{verify}. AgentOS supplies object, region, or target references, and the perception backend returns detected instances, confidence scores, view evidence, target-verification results, and memory-update triggers. Navigation and manipulation skills use this evidence for target selection and outcome verification, while the memory layer stores it as object observations, view descriptors, or refreshed scene-graph evidence.

\subsection{Spatial Navigation and Exploration: \holonavi}
\label{subsec:holonavi}
\begin{figure}[t]
    \vspace{-4pt}    
    \centering
    \begin{overpic}[width=0.95\textwidth, keepaspectratio]{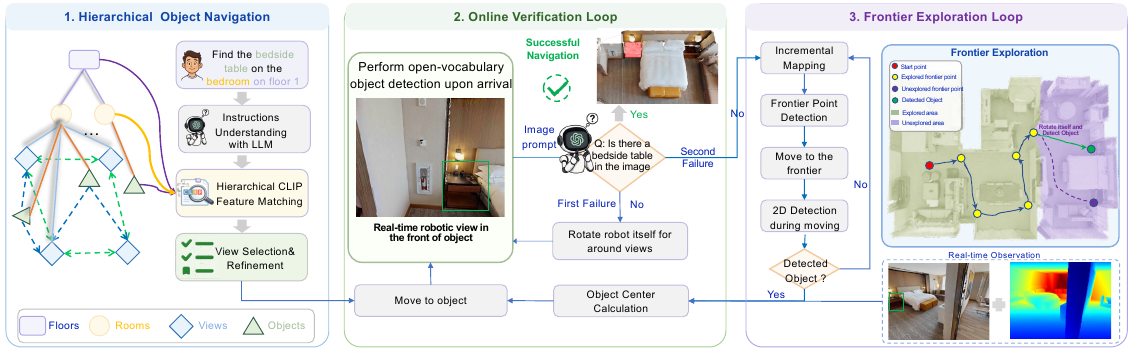}
    \end{overpic}
    \caption{
    \textbf{\holonavi object-navigation pipeline.}
    Given a language goal, \holonavi first performs \textbf{hierarchical semantic navigation} by parsing the instruction and matching it against floor-, room-, view-, and object-level memory.
    After reaching a candidate viewpoint, an \textbf{online verification loop} validates the target with open-vocabulary detection and local viewpoint adjustment.
    If verification fails, a \textbf{frontier exploration loop} expands the map with continuous mapping and online object detection until the target is found or the search is exhausted.
    }
    \label{fig:nav_reasoning_pipeline}  
    \vspace{-4pt}
\end{figure}

\holonavi implements the spatial-navigation skill family in \modelname. Its inputs include language-level destinations, object goals, frontier targets, and target poses, while its status outputs report goal reachability, localization confidence, blocked routes, verification failures, and exploration progress. Internally, \holonavi turns these skill calls into executable navigation targets through hierarchical object navigation, online verification, and frontier exploration.

\textbf{\calmfont{Hierarchical object navigation.}} Given a \texttt{move\_to} or \texttt{find} skill call, an LLM first parses the request into structured spatial and semantic queries, such as floor, room, and object targets. \holonavi then performs fast matching over the HMSG to retrieve candidate rooms, views, and object instances. This fast stage uses hierarchical CLIP-feature matching to prune the search space before expensive visual reasoning.

\textbf{\calmfont{Online verification loop.}} Fast matching can return visually similar but incorrect candidates, especially for small objects, ambiguous room names, or stale memory. To refine the target, \holonavi sends candidate goal views to a VLM for visual verification and slow reasoning, following the fast-to-slow reasoning design of FSR-VLN~\citep{fsr_vln}. If the VLM confirms the target object in the selected view, \holonavi computes the object center and dispatches navigation to the verified goal. If verification fails, the system rotates the robot to collect surrounding views, re-runs object detection and VLM verification, and reports the failed candidate so that AgentOS can update memory or select a different sub-goal before committing to navigation.

\textbf{\calmfont{Active spatial exploration.}} When the current HMSG and online views cannot localize the target, AgentOS triggers an \emph{active exploration} skill instead of planning from missing state. The trigger can come from unresolved room or object references, low-confidence scene-graph retrieval, frontier regions in the metric map, or failed online verification caused by stale memory.

The exploration skill scores candidate viewpoints by expected information gain, semantic relevance to the task, traversability, and safety constraints, then sends the selected viewpoint or frontier route to \holonavi for execution. After each exploration step, the skill reports newly observed regions, unresolved targets, and traversal failures to AgentOS. The memory update layer then fuses these events into geometry memory, semantic instances, and the affected HMSG subgraph, allowing AgentOS to query expanded spatial memory and resume the original task with better grounding.

\subsection{Manipulation: \holobrain}
\label{subsec:holobrain}

\holobrain is the manipulation VLA backend used by \modelname~\citep{holobrain0}. In the skill layer, \holobrain implements manipulation calls such as \texttt{pick}, \texttt{place}, \texttt{open}, \texttt{handover}, \texttt{push}, and \texttt{fold}. Each call combines task intent from AgentOS, current visual observations, robot embodiment priors, and optional object grounding from semantic memory. Given this structured context, \holobrain performs end-to-end policy inference and outputs executable arm, gripper, or dual-arm actions for object-level interaction.

As a monitored skill backend, \holobrain reports execution evidence rather than only returning a terminal success flag. 
The status stream includes object-not-found events, object motion, grasp failure, unreachable poses, collision risk, low policy confidence, and user-confirmation requirements. 
These signals make VLA execution usable by the AgentOS runtime: failures become explicit planning evidence for retries, perception updates, clarification, or re-planning.

For long-horizon mobile manipulation, \modelname composes \holobrain with navigation and perception skills through the same command/status interface. 
\holonavi first brings the robot to a task-relevant region or object viewpoint; \holobrain then executes the local manipulation skill from the current observation and target context. 
If the object is absent, occluded, unreachable, or requires a different approach pose, the reported status can trigger memory retrieval, active exploration, navigation repositioning, or a new manipulation sub-plan. 
This loop provides a practical interface for composing VLA-based local control with task decomposition and navigation in long-horizon manipulation demonstrations.

\subsection{Whole-Body Motion: \holomotion}
\label{subsec:holomotion}

\holomotion provides the humanoid whole-body motion backend in \modelname~\citep{holomotion1}. It exposes reference tracking, velocity control, posture adjustment, and recovery skills, while reporting progress, balance state, contact risk, velocity error, and recovery availability. This interface lets AgentOS request high-level motion goals while the backend handles low-level joint control.

\holomotion supports two execution modes. In motion-tracking mode, the humanoid follows retargeted demonstrations for interaction behaviors such as waving, bowing, handshaking, or dancing. In velocity-tracking mode, it follows commanded linear and angular velocities for locomotion skills such as walking, turning, stopping, and recovery. These skills extend \modelname from navigation and manipulation planning to physically grounded humanoid behavior.

Through the same command/status interface, \holomotion composes with other skill families. It realizes \holonavi outputs as stable walking and turning commands, and it supports \holobrain by adjusting torso pose, reaching a stable interaction posture, or recovering after contact failure. AgentOS uses the reported motion state to continue, slow down, retry, trigger recovery, or re-plan when the humanoid state no longer supports the current task.

\subsection{Cross-Embodiment Coordination through Shared Interfaces}
\label{subsec:cross_embodiment}

The skill execution interface also enables lightweight cross-embodiment coordination. Heterogeneous robots share memory records, typed skill calls, and status events, so AgentOS can assign different parts of one task to different bodies without exposing their low-level controllers.

All embodiments write observations, detections, map updates, and skill outcomes into the same hierarchical 3D memory, with each update tagged by spatial evidence, temporal context, and the reporting embodiment. AgentOS can therefore reuse discoveries made by one robot when planning for another, such as using a mobile base to populate candidate object locations before a humanoid retrieves them for manipulation.

AgentOS binds each task-level skill call to an embodiment according to capability, location, availability, and safety state. During execution, shared status events report pose, sensor health, active sub-goals, progress, failure modes, and recoverability, allowing AgentOS to coordinate concurrent execution, avoid workspace conflicts, and transfer responsibility when a robot becomes blocked. This mechanism composes robots through shared memory, typed skill calls, and observable status events rather than a separate controller.


\section{Memory Layer: Spatial and Temporal Memory for Embodied Agents}
\label{sec:memory}

The Memory Layer provides spatial and temporal memory for \modelname. Spatial memory converts sensor streams into a persistent 3D world representation that Embodied AgentOS can query for grounding, localization, navigation, and manipulation. It includes geometry, topology, and occupancy; robot pose and localization; an open-vocabulary 3D semantic map; and a hierarchical multimodal scene graph (HMSG) for room-, view-, and object-level reasoning. Temporal memory records the evolving goal and plan state, execution and recovery traces, and outcome summaries produced during the AgentOS loop. For a task query, the memory layer returns candidate places, objects, poses, task history, and recent execution evidence. After execution, it updates the affected spatial records and temporal traces.

\subsection{Spatial Memory: Metric Geometry, Topology, and Localization}
\label{subsec:geom_recon}

Geometry memory provides the metric substrate for embodied skills. It maintains the coordinate frame, robot pose, dense geometry, traversability evidence, and localization index used by semantic mapping, navigation, manipulation preparation, and memory update. To decouple downstream AgentOS behavior from the sensor configuration, \modelname exposes a unified geometry interface that can be populated by either a LiDAR-based backend or a vision-only backend.

\textbf{\calmfont{LiDAR-based backend.}} The LiDAR-based backend fuses LiDAR, IMU, and camera observations into a metric point cloud or mesh, following tightly coupled LiDAR-inertial-visual odometry systems such as FAST-LIVO~\citep{fast_livo}. This backend provides a robust reference map for safety-critical deployment, controlled evaluation, and sensor calibration.

\textbf{\calmfont{Vision-only backend.}} The vision-only backend builds geometry memory from synchronized RGB streams with GeoFlow-SLAM++, a multi-camera extension of GeoFlow-SLAM~\citep{geoflow_slam}. Given calibrated cameras $\{I_{c_k}\}_{k=1}^{N}$ with $N \geq 3$, a 3D foundation model $\Phi_\theta$ predicts dense depth from multi-view images. The backend back-projects the predicted depths and aligns them into the robot body frame through known camera extrinsics, producing a shared local geometric structure without requiring physical RGB-D depth hardware.

GeoFlow-SLAM++ improves robustness by using multi-camera geometry for tracking, mapping, and relocalization. During tracking, a multi-view two-stage optical-flow strategy maintains feature continuity and unifies camera matches in the body frame for joint pose estimation. During local mapping, the backend jointly optimizes visual reprojection, point-plane, and IMU constraints to refine the robot state in occluded, low-texture, or corridor-like scenes. For relocalization, it loads a pre-built map atlas, aggregates Bag-of-Words (BoW) vectors from all cameras into a unified retrieval query, and refines candidate keyframes through cross-view 3D-to-2D correspondences and joint bundle adjustment. Once relocalized, the robot can perform pure localization on the existing map.


\subsection{Semantic Memory: Open-Vocabulary 3D Semantic Mapping}
\label{subsec:semantic_mapping}
\begin{figure}[ht]
\centering
\includegraphics[width=0.9\textwidth]{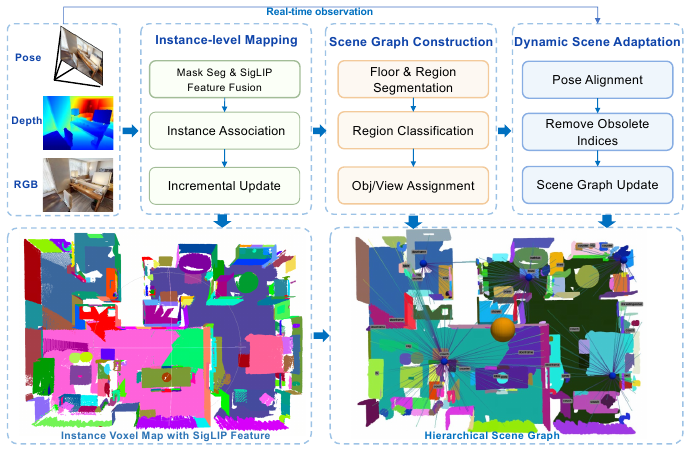}
\caption{\textbf{Overview of the \modelname semantic mapping framework.} The semantic memory lifts open-vocabulary 2D features onto geometry memory, associates observations with persistent 3D instances, and provides language-queryable object and region evidence for AgentOS.}
\label{fig:mapping_framework}
\vspace{-3mm}
\end{figure}

Semantic memory turns metric geometry into language-grounded 3D spatial memory. On top of geometry memory, an open-vocabulary online mapping module lifts 2D foundation-model features onto 3D points, voxels, and instances, following the broader direction of open-vocabulary online semantic mapping and semantic SLAM with unified geometry-instance representations~\citep{ovo_mapping,iris_slam}. This representation allows Embodied AgentOS to retrieve objects, regions, and visual evidence with natural-language queries.

For each 2D mask produced by SAM2, we compute three SigLIP descriptors: $\mathbf{d}_0$ for the full keyframe, $\mathbf{d}_1$ for the masked segment, and $\mathbf{d}_2$ for the minimum bounding box that contains the segment. We merge these descriptors with a per-dimension weighted average,
\begin{equation}
    \mathbf{d} = \sum_{i=0}^{2} \mathbf{w}_i \odot \mathbf{d}_i,
\end{equation}
where $\mathbf{w}_i \in \mathbb{R}^d$ and $\odot$ denotes the Hadamard product. The fused descriptor preserves global context, segment-specific appearance, and local object detail for open-vocabulary retrieval.
    
To maintain persistent object identities, semantic memory associates new 2D observations with existing 3D instances. We first project the existing 3D instances $\mathcal{V}_{t-1}$ into the current camera view to obtain projected masks $\{\tilde{m}_j\}$. We then compute $\mathrm{IoU}(m_k, \tilde{m}_j)$ between current masks $\{m_k\}$ and projected masks, and merge matched observations into the corresponding 3D instances. Confident observations without matches initialize new persistent 3D instances.
\begin{figure*}[ht]
    \vspace{-6pt}
    \centering
    \includegraphics[width=1.0\linewidth]{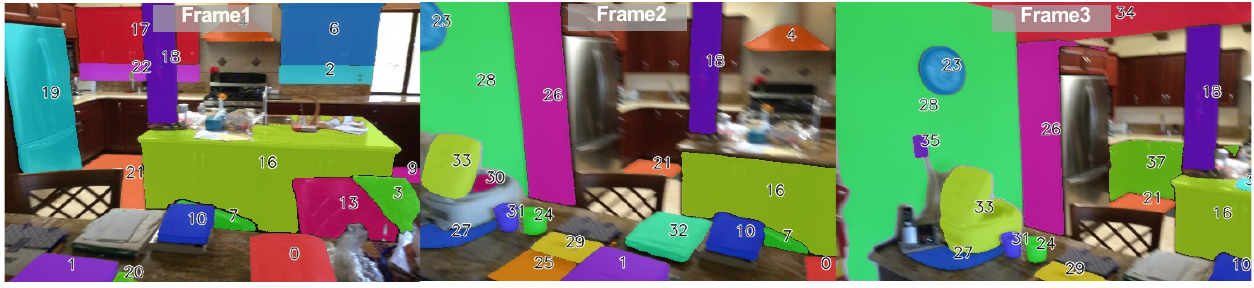}
    \caption{\textbf{Instance association.} The semantic memory projects existing 3D instances into the current view and matches them with new 2D masks to maintain persistent object identities.}
    \label{fig:instance_association}
    \vspace{-6pt}
    \end{figure*}

\subsection{Hierarchical Multimodal Scene Graph for Structured Spatial Retrieval}
\label{subsec:hmsg}

The hierarchical multimodal scene graph (HMSG) provides the structured retrieval interface between semantic memory and Embodied AgentOS. Following FSR-VLN~\citep{fsr_vln}, HMSG organizes memory into floor, room, view, and object layers, with hierarchical edges for spatial containment and topological edges for view connectivity and object visibility. In \modelname, we reuse this hierarchy as a persistent memory index for task planning, target grounding, and recovery.

\begin{figure}[t]
    \centering
    \includegraphics[width=1.0\linewidth]{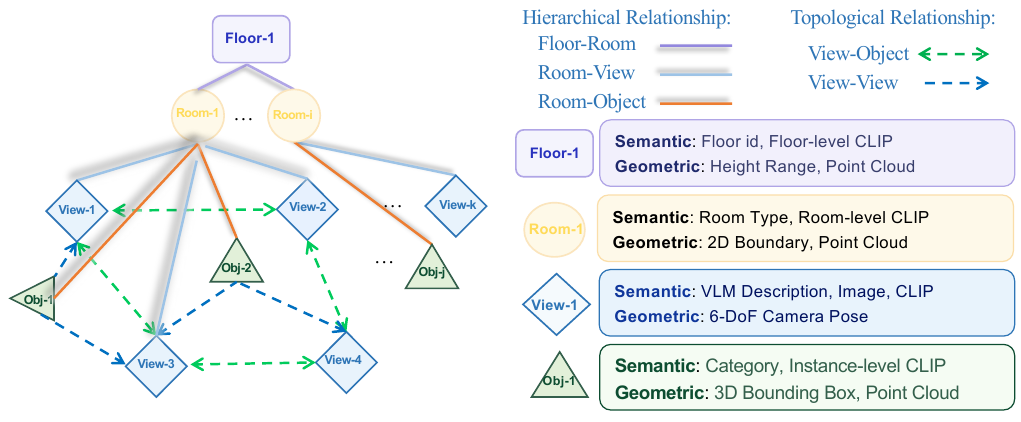}
    \caption{\textbf{Hierarchical Multimodal Scene Graph (HMSG).}
    HMSG organizes memory into floor, room, view, and object layers. In \modelname, this hierarchy acts as the retrieval index for AgentOS, supporting coarse-to-fine target grounding, VLM-based verification, and memory updates from execution feedback.}
    \label{fig:hmsg}
    \vspace{-6pt}
\end{figure}

As shown in Fig.~\ref{fig:hmsg}, each layer provides a different retrieval granularity. Floor and room nodes narrow a query to a coarse spatial region; view nodes connect robot poses with visual evidence; and object nodes provide persistent instance-level targets for navigation or manipulation. This hierarchy avoids exhaustive search over raw geometry or all detected instances while preserving the visual context needed for downstream VLM-based target verification.

The view layer bridges geometric memory and visual reasoning. Conventional floor-room-object hierarchies provide useful spatial structure but often rely on direct feature matching for target localization. Image-based topological graphs provide visual evidence but are less grounded in room- and object-level memory. HMSG keeps both forms of evidence: view nodes store candidate perspectives and visibility links to objects, enabling AgentOS to prune the search space geometrically before verifying a small set of candidate views with a VLM.

During closed-loop execution, HMSG is queried before skill dispatch and refreshed after memory updates. Failed verification, newly explored views, and skill outcomes can change the candidate evidence available for later planning. The detailed update procedure is described in Sec.~\ref{subsec:dynamic_memory_update}.

\subsection{Temporal Memory: Task State, Execution Trace, and Experience}
\label{subsec:temporal_memory}

Temporal memory complements spatial memory by preserving what the agent is trying to do, what has already happened, and which recovery attempts have been made. Spatial memory answers where objects, rooms, poses, and traversable regions are; temporal memory answers which goal is active, which plan steps remain unresolved, which skills succeeded or failed, and what evidence should guide AgentOS.

\textbf{\calmfont{Goal and plan state.}} AgentOS stores the parsed user intent, active skill graph, pending sub-goals, assigned embodiments, preconditions, and unresolved references as temporal state. This state allows the runtime to resume a long-horizon task after interruptions, bind later skill calls to earlier decisions, and avoid losing context when execution spans multiple perception and action cycles.

\textbf{\calmfont{Execution and recovery trace.}} Each dispatched skill appends command parameters, retrieved context, status events, verification results, failure modes, retries, user clarifications, and memory-update triggers to the temporal trace. These records give AgentOS explicit evidence for deciding whether to continue, retry, re-plan, ask for clarification, or mark a failure as unrecoverable.

\textbf{\calmfont{Outcome and experience summary.}} After a task segment terminates, the memory layer stores a compact outcome summary that records the final state, successful or failed recovery strategy, changed objects or locations, and user-facing explanation. These summaries support later planning and debugging by turning raw execution logs into queryable experience rather than only preserving low-level ROS2 traces.

\subsection{Memory Update: Updating Spatial Records and Temporal Traces}
\label{subsec:dynamic_memory_update}

The memory layer is not a one-shot scan; it is a stateful representation that changes with robot execution. Three event types trigger memory updates: new sensor observations that conflict with existing memory, skill outcomes that move or remove objects, and explicit user feedback that corrects the current state. Embodied AgentOS uses these updates to keep later planning steps consistent with both the physical world and the current task history.

Given a new observation, \modelname first relocalizes the robot in geometry memory and then updates the local metric map around the current view. The update procedure removes or refreshes points and voxels that conflict with the new depth and color evidence while fusing newly observed geometry into local memory. Semantic memory then associates new masks with existing instances or creates new instances when no reliable match exists. Finally, HMSG refreshes only the affected subgraph by recomputing changed objects, their parent room nodes, visible view nodes, and local spatial relations without rebuilding the full graph.

Skill outcomes provide a second update path. A successful \texttt{pick} can mark an object as carried or absent from its previous support surface; a failed \texttt{move\_to} can attach blockage or localization-uncertainty evidence to a route; a user correction can rename an object or room. In parallel, temporal memory appends the corresponding command, status, verification result, and recovery decision. The memory layer stores these spatial and temporal updates as queryable entries, allowing AgentOS to retrieve the latest world state and task history in later planning loops.

\section{Experiments}
\label{sec:experiments}

\subsection{Experimental Setup}
\label{subsec:exp_setup}

Our evaluation separates repeatable quantitative measurements from broader real-robot system demonstrations. The quantitative experiments cover two components that can be tested under fixed protocols: long-horizon navigation, where the full AgentOS navigation loop combines spatial memory, goal verification, and execution feedback; and 3D semantic mapping, where the memory layer provides language-queryable spatial grounding. The qualitative experiments then illustrate how the same memory, skill, and monitoring interfaces compose heterogeneous robot behaviors, including humanoid motion, object search, cross-robot coordination, and mobile manipulation.

\textbf{\calmfont{Platforms and environments.}} We deploy \modelname on three real robot platforms: the Unitree G1 humanoid, the R1 humanoid, and a wheeled dual-arm mobile manipulator. The humanoid platforms use the memory-layer sensing stack, including RGB-D cameras, audio interaction devices, and onboard computing. The mobile manipulator uses a wheeled base, two 6-DoF arms with grippers, an elevated RGB-D camera for global perception, and onboard computing and control modules. We also use repeatable simulated scenes for navigation debugging and controlled comparison. Across simulation and hardware, AgentOS uses the same command/status interface, so skill dispatch, monitoring, and task-state updates follow a consistent execution protocol across platforms.

\textbf{\calmfont{Metrics and evidence scope.}} For 3D semantic mapping, we report open-vocabulary semantic segmentation metrics, including mIoU, mAcc, frequency-weighted mIoU, and frequency-weighted accuracy. For long-horizon navigation, we report success rate (SR) and success weighted by path length (SPL), as well as real-robot goal-reaching success under Top-1 and Top-5 candidate selection. For full-stack embodied-agent behavior, we report qualitative execution traces rather than a single end-to-end success benchmark, because manipulation, whole-body motion, and cross-embodiment coordination use heterogeneous hardware and are not yet standardized under one repeatable protocol.
\begin{figure}[t]
    \centering
    \includegraphics[width=0.8\linewidth]{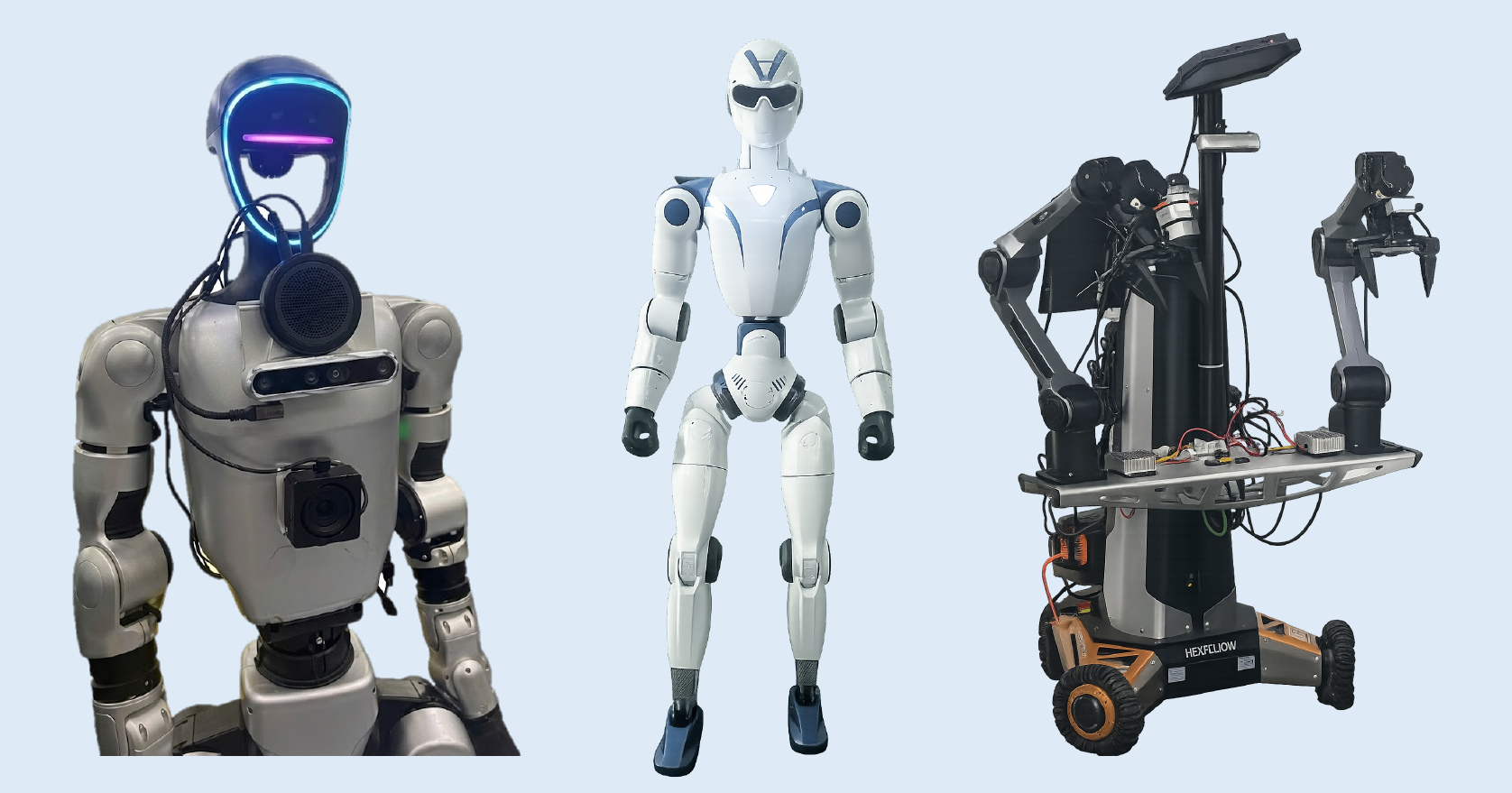}
    \caption{\textbf{Robot platforms used in real-world experiments.}
    \modelname is deployed on heterogeneous embodiments, including humanoid platforms for navigation, interaction, and whole-body motion, and a wheeled dual-arm platform for mobile manipulation.}
    \label{fig:exp_setting}
    \vspace{-6pt}
\end{figure}

\subsection{Long-Horizon Navigation: Agent-Loop Evaluation}
\label{subsec:exp_navi}

We evaluate whether the AgentOS navigation loop improves long-horizon goal reaching beyond a standalone navigation backend. The evaluation uses two complementary settings. In simulation, we follow the zero-shot ObjectNav setting used by MSGNav~\citep{msgnav}, which stresses open-vocabulary goal grounding, spatial reasoning, and path efficiency without hardware noise. On real robots, we evaluate \holoagentnav in physical apartments with matched digital twins. This pairing separates simulation-level comparison with prior navigation systems from real-world transfer under localization, actuation, and perception noise.

\textbf{\calmfont{Simulation evaluation.}} The simulated benchmark follows the HM3D-ObjNav protocol adopted by MSGNav~\citep{msgnav}. Each episode provides a goal object category and a start pose in an unseen indoor scene; the agent must actively explore until it reaches a valid target viewpoint. We report Success Rate (SR) and Success weighted by Path Length (SPL), following the standard definitions used in MSGNav. SR measures whether the agent reaches the goal, while SPL additionally penalizes unnecessarily long successful paths. Because the simulator abstracts away low-level robot execution, this setting primarily evaluates semantic grounding, spatial-memory retrieval, viewpoint selection, and route efficiency.

The comparison set contains two groups. The first group covers representative zero-shot navigation systems reported under the HM3D-ObjNav setting, including VLFM~\citep{vlfm}, SG-Nav~\citep{sg_nav}, DORAEMON~\citep{doraemon_nav}, WMNav~\citep{wmnav}, and MSGNav~\citep{msgnav}. These methods use frontier maps, scene graphs, world models, or multimodal memory structures for goal grounding. The second group contains the fast-matching and slow-reasoning variants of FSR-VLN~\citep{fsr_vln}, which we include as reference hierarchical-memory baselines for isolating the effect of wrapping the navigation backend with AgentOS. \holoagentnav denotes the full navigation stack with memory retrieval, candidate verification, skill execution, monitoring, and feedback-driven state updates.

\begin{table}[ht]
\centering
\caption{\textbf{Experiments on the HM3D-ObjNav benchmark.} The simulated setting follows the evaluation protocol of MSGNav~\citep{msgnav}. FSR-VLN variants are included as reference hierarchical-memory baselines, while \holoagentnav denotes the AgentOS-wrapped navigation stack.}
\label{tab:fsrvln_sim}
\begin{tabular}{lcc}
\toprule
Method & SR (\%) $\uparrow$ & SPL (\%) $\uparrow$ \\ 
\midrule
SG-Nav~\citep{sg_nav} & 49.6 & 25.5 \\ 
VLFM~\citep{vlfm} & 62.6 & 31.0 \\ 
DORAEMON~\citep{doraemon_nav} & 66.5 & 20.6 \\ 
WMNav~\citep{wmnav} & 72.2 & 33.3 \\ 
MSGNav~\citep{msgnav} & 74.1 & 33.4 \\ 
FSR-VLN (fast-matching)~\citep{fsr_vln} & 72.1 & 36.9 \\ 
FSR-VLN (slow-reasoning)~\citep{fsr_vln} & 80.8 & 41.0 \\ 
\holoagentnav (w AgentOS loop) & \textbf{82.6} & \textbf{42.8} \\ 
\bottomrule
\end{tabular}
\end{table}

\textbf{\calmfont{Real-robot evaluation.}} The real-robot evaluation follows the benchmark construction of FSR-VLN~\citep{fsr_vln} and tests whether \holoagentnav reaches language-specified goals in physical apartments. We evaluate goal reaching under Top-1 and Top-5 candidate selection with 1.0\,m, 2.0\,m, and 3.0\,m distance thresholds, separating strict goal localization from more tolerant candidate retrieval. This setting tests both the memory query that proposes navigable goals and the closed-loop execution that reaches those goals on the robot.

\begin{table*}[ht]
\centering
\caption{\textbf{Real-robot navigation evaluation.} We report success rate under Top-1 and Top-5 goal candidates with 1.0\,m, 2.0\,m, and 3.0\,m distance thresholds.}
\label{tab:fsrvln_real}
\resizebox{\textwidth}{!}{%
\begin{tabular}{lcccccc}
\toprule
Method & Top-1@1.0m  & Top-1@2.0m  & Top-1@3.0m  & Top-5@1.0m  & Top-5@2.0m  & Top-5@3.0m  \\ 
\midrule
OK-Robot~\citep{ok_robot} & 60.92 & 60.92 & 60.92 & 63.22 & 63.22 & 63.22 \\ 
MobilityVLA~\citep{mobility_vla} & 34.48 & 59.77 & 75.86 & -- & -- & -- \\ 
HOV-SG~\citep{hovsg} & 51.72 & 57.47 & 58.62 & 77.00 & 81.61 & 82.76 \\ 
FSR-VLN~\citep{fsr_vln} & 91.95  & 91.95  & 94.25  & 94.25  & 96.55  & 96.55 \\ 
\holoagentnav & \textbf{97.70} & \textbf{97.70} & \textbf{97.70} & \textbf{98.90} & \textbf{98.90} & \textbf{98.90} \\ 
\bottomrule
\end{tabular}%
}
\end{table*}

\textbf{\calmfont{Discussion.}} \holoagentnav improves both simulated path efficiency and real-robot goal reaching. In simulation, it improves over FSR-VLN and outperforms the strongest published HM3D-ObjNav baseline, MSGNav, on both SR and SPL, indicating that feedback-driven execution preserves path efficiency rather than merely increasing success. On real robots, the identical scores across 1.0\,m, 2.0\,m, and 3.0\,m thresholds show that successful trials already stop within the strict threshold, while the remaining failures require better recovery rather than relaxed localization. These results support AgentOS as a practical way to connect spatial memory, goal verification, monitored execution, and feedback-driven recovery.


\subsection{Memory Evaluation: Semantic Mapping and Update Visualization}
\label{subsec:exp_semantic_mapping}

We evaluate whether the memory layer provides useful language-grounded 3D spatial memory for downstream planning. This experiment isolates perception and memory from long-horizon execution, so the results assess whether \modelname can lift 2D open-vocabulary evidence into persistent 3D objects, regions, and scene-graph nodes. A useful memory layer should preserve semantic accuracy under zero-shot queries and maintain stable object identities as the robot revisits the environment.

\textbf{\calmfont{3D semantic mapping evaluation.}} We evaluate open-vocabulary 3D semantic mapping on ScanNet and Replica using mIoU, mAcc, frequency-weighted mIoU, and frequency-weighted accuracy. The first two metrics measure category-level semantic quality, while the frequency-weighted metrics emphasize common room-scale entities that are often queried during navigation and manipulation. We compare against representative offline and online semantic mapping systems to separate raw mapping quality from deployability in an embodied-agent loop.

\begin{table}[ht]
\centering
\caption{\textbf{Zero-shot semantic mapping on ScanNet and Replica.} We compare offline and online mapping methods using category-level and frequency-weighted metrics. Metrics are normalized \textbf{column-wise} from \colorbox{green1}{green} (best) to \colorbox{red2}{red} (worst).}
\label{tab:semantic_eval_combined}

\begin{subtable}{0.48\textwidth}
    \centering
    \caption{ScanNet dataset}
    \resizebox{\linewidth}{!}{%
    \setlength{\tabcolsep}{4pt}
    \begin{tabular}{l c cccc}
    \toprule
    \textbf{Method} & \textbf{Online} & \textbf{mIoU}$\uparrow$ & \textbf{mAcc}$\uparrow$ & \textbf{f-mIoU}$\uparrow$ & \textbf{f-Acc}$\uparrow$ \\
    \midrule
    Open-Gaussian & \textcolor{red}{\xmark} & \getsem{08.64}{06.69}{39.93}08.64 & \getsem{17.86}{14.89}{55.36}17.86 & \getsem{23.71}{16.62}{53.62}23.71 & \getsem{26.44}{19.57}{66.78}26.44 \\
    
    Lang-Splat & \textcolor{red}{\xmark} & \getsem{07.22}{06.69}{39.93}07.22 & \getsem{21.01}{14.89}{55.36}21.01 & \getsem{27.59}{16.62}{53.62}27.59 & \getsem{32.21}{19.57}{66.78}32.21 \\
    
    Grasp-Splats & \textcolor{green}{\cmark} & \getsem{06.69}{06.69}{39.93}06.69 & \getsem{14.89}{14.89}{55.36}14.89 & \getsem{16.62}{16.62}{53.62}16.62 & \getsem{19.57}{19.57}{66.78}19.57 \\
    
    Omni-Map & \textcolor{green}{\cmark} & \getsem{25.42}{06.69}{39.93}25.42 & \getsem{50.93}{14.89}{55.36}50.93 & \getsem{50.86}{16.62}{53.62}50.86 & \getsem{57.05}{19.57}{66.78}57.05 \\
    
    HOV-SG & \textcolor{red}{\xmark} & \getsem{20.76}{06.69}{39.93}20.76 & \getsem{41.50}{14.89}{55.36}41.50 & \getsem{38.34}{16.62}{53.62}38.34 & \getsem{45.50}{19.57}{66.78}45.50 \\
    
    Concept-Fusion & \textcolor{green}{\cmark} & \getsem{08.50}{06.69}{39.93}08.50 & \getsem{31.81}{14.89}{55.36}31.81 & \getsem{30.05}{16.62}{53.62}30.05 & \getsem{36.64}{19.57}{66.78}36.64 \\
    
    Concept-Graph & \textcolor{green}{\cmark} & \getsem{16.29}{06.69}{39.93}16.29 & \getsem{34.07}{14.89}{55.36}34.07 & \getsem{33.29}{16.62}{53.62}33.29 & \getsem{41.60}{19.57}{66.78}41.60 \\
    
    Open-Fusion & \textcolor{green}{\cmark} & \getsem{18.02}{06.69}{39.93}18.02 & \getsem{44.31}{14.89}{55.36}44.31 & \getsem{43.82}{16.62}{53.62}43.82 & \getsem{51.09}{19.57}{66.78}51.09 \\
    
    HoloAgent-Memory & \textcolor{green}{\cmark} & \getsem{31.58}{06.69}{39.93}31.58 & \getsem{45.54}{14.89}{55.36}45.54 & \getsem{47.43}{16.62}{53.62}47.43 & \getsem{61.58}{19.57}{66.78}61.58 \\
    
    \bottomrule
    \end{tabular}%
    }
\end{subtable}
\hfill
\begin{subtable}{0.48\textwidth}
    \centering
    \caption{Replica dataset}
    \resizebox{\linewidth}{!}{%
    \setlength{\tabcolsep}{4pt}
    \begin{tabular}{l c cccc}
    \toprule
    \textbf{Method} & \textbf{Online} & \textbf{mIoU}$\uparrow$ & \textbf{mAcc}$\uparrow$ & \textbf{f-mIoU}$\uparrow$ & \textbf{f-Acc}$\uparrow$ \\
    \midrule
    Open-Gaussian & \textcolor{red}{\xmark} & \getsem{06.82}{04.75}{31.15}06.82 & \getsem{16.66}{16.66}{44.54}16.66 & \getsem{15.41}{15.41}{64.42}15.41 & \getsem{18.08}{18.08}{72.30}18.08 \\
    
    Lang-Splat & \textcolor{red}{\xmark} & \getsem{10.00}{04.75}{31.15}10.00 & \getsem{22.93}{16.66}{44.54}22.93 & \getsem{39.69}{15.41}{64.42}39.69 & \getsem{44.16}{18.08}{72.30}44.16 \\
    
    Grasp-Splats & \textcolor{green}{\cmark} & \getsem{10.42}{04.75}{31.15}10.42 & \getsem{23.79}{16.66}{44.54}23.79 & \getsem{42.67}{15.41}{64.42}42.67 & \getsem{52.39}{18.08}{72.30}52.39 \\
    
    Omni-Map & \textcolor{green}{\cmark} & \getsem{29.06}{04.75}{31.15}29.06 & \getsem{44.54}{16.66}{44.54}44.54 & \getsem{64.42}{15.41}{64.42}64.42 & \getsem{72.22}{18.08}{72.30}72.22 \\
    
    HOV-SG & \textcolor{red}{\xmark} & \getsem{23.79}{04.75}{31.15}23.79 & \getsem{39.59}{16.66}{44.54}39.59 & \getsem{48.86}{15.41}{64.42}48.86 & \getsem{55.15}{18.08}{72.30}55.15 \\
    
    Concept-Fusion & \textcolor{green}{\cmark} & \getsem{04.75}{04.75}{31.15}04.75 & \getsem{19.29}{16.66}{44.54}19.29 & \getsem{25.30}{15.41}{64.42}25.30 & \getsem{28.99}{18.08}{72.30}28.99 \\
    
    Concept-Graph & \textcolor{green}{\cmark} & \getsem{16.46}{04.75}{31.15}16.46 & \getsem{31.51}{16.66}{44.54}31.51 & \getsem{35.69}{15.41}{64.42}35.69 & \getsem{42.44}{18.08}{72.30}42.44 \\
    
    Open-Fusion & \textcolor{green}{\cmark} & \getsem{16.37}{04.75}{31.15}16.37 & \getsem{35.15}{16.66}{44.54}35.15 & \getsem{51.65}{15.41}{64.42}51.65 & \getsem{60.37}{18.08}{72.30}60.37 \\
    
    HoloAgent-Memory & \textcolor{green}{\cmark} & \getsem{29.93}{04.75}{31.15}29.93 & \getsem{43.60}{16.66}{44.54}43.60 & \getsem{57.00}{15.41}{64.42}57.00 & \getsem{65.39}{18.08}{72.30}65.39 \\
    
    \bottomrule
    \end{tabular}%
    }
\end{subtable}

\vspace{-6pt}
\end{table}

\textbf{\calmfont{Discussion.}} Table~\ref{tab:semantic_eval_combined} shows that the memory layer provides competitive zero-shot semantic grounding while retaining online operation. It performs strongly on ScanNet and remains competitive among online methods on Replica, supporting semantic memory as a queryable interface for AgentOS. The remaining gap on Replica, especially for frequency-weighted metrics, suggests that large-area entities and multi-view feature aggregation remain important directions for refinement.

\textbf{\calmfont{Memory update visualization.}} Quantitative segmentation metrics do not show whether object identities remain stable under repeated observations or environmental changes. We therefore visualize two memory operations required by the AgentOS loop: instance association, which merges new 2D observations into persistent 3D objects, and dynamic update, which refreshes changed regions without rebuilding the whole map. Together, these cases show how the memory layer keeps object-level evidence queryable as new observations arrive during execution.


\begin{figure}[ht]
\centering
\includegraphics[width=0.8\textwidth]{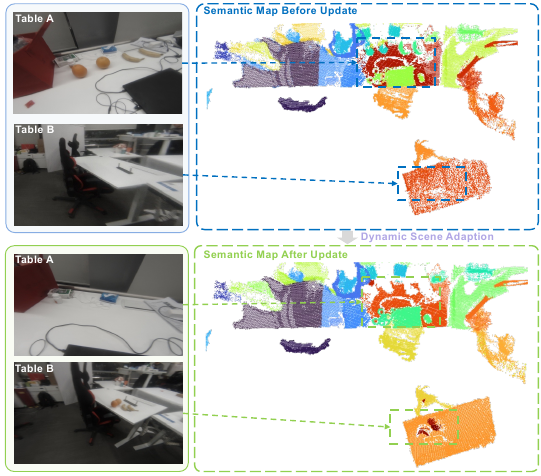}
\caption{\textbf{Dynamic memory update.} The memory layer localizes the robot in the existing map, refreshes changed geometry and semantic instances around the current observation, and updates only the affected scene-graph nodes and relations.}
\label{fig:dynamic_update}
\end{figure}

\subsection{Qualitative Closed-Loop Robot Execution}
\label{subsec:exp_qualitative}

Beyond scalar navigation and mapping metrics, we visualize representative real-robot executions that exercise the full \modelname runtime. Fig.~\ref{fig:framework_vis} covers humanoid motion, object search, cross-robot coordination, and mobile manipulation. These cases are not a standardized benchmark; instead, they illustrate how AgentOS composes heterogeneous skills through shared task state, status feedback, and memory-grounded planning.

The traces highlight the role of the typed skill abstraction. AgentOS can plan over spatial memory, verify targets through navigation and perception feedback, coordinate multiple robots through a shared command/status interface, and decompose long-horizon tasks such as laundry folding into navigation, perception, placement, and manipulation steps. A rigorous end-to-end benchmark for manipulation, whole-body motion, and cross-embodiment collaboration remains future work.

\section{Related Work}
\label{sec:related}

\paragraph{\calmfont{From Digital to Embodied Agent Frameworks.}}
Digital LLM agents have converged on a practical execution loop: they reason over structured state, invoke tools, inspect feedback, store memory, and revise subsequent actions~\citep{react,toolformer,reflexion,generative_agents,autogen,voyager}. This loop works well in software settings because tools expose clean inputs and outputs, and observations arrive as text, code, web pages, or API responses. Physical embodiment weakens these assumptions: robot actions unfold over time, depend on embodiment-specific controllers and uncertain 3D state, and may fail partially or unsafely. An embodied agent framework therefore requires system interfaces that digital frameworks rarely provide, including persistent metric-semantic memory, typed access to heterogeneous robot skills, and runtime monitoring for failure detection and re-planning. Existing robot middleware and recent LLM-enabled robot runtimes provide important pieces of this stack. These pieces range from typed communication and modular deployment in ROS2 to language-level task interpretation, ROS action or service generation, cross-embodiment skill libraries, shared state, and modular capability packages~\citep{ros2,ros_llm,roboos,emos,aeros}. \modelname follows this transition from digital to embodied agents, but centers the framework on a closed-loop physical execution interface that combines instruction-conditioned policies for physical behavior with persistent 3D scene understanding for large-scale grounding.

\paragraph{\calmfont{Instruction-Conditioned Robot Policies.}}
Instruction-conditioned robot policies provide the behavior layer for embodied agents by mapping language-level goals to executable physical actions. Task-and-motion planning couples symbolic task structure with geometric feasibility~\citep{kaelbling2011hierarchical,garrett2021integrated}, while language-conditioned robot systems use language models to select, sequence, or synthesize executable robot behaviors~\citep{saycan,inner_monologue,code_as_policies}. Recent embodied multimodal models, vision-language-action policies, video-action models, and world-action models,
further improve generalization by learning from large-scale visual, language, video, and robot-action data~\citep{palm_e,rt2,kim2024openvla,robovlm,bjorck2025gr00t,pi0,pi05,pi07,gr1,lingbotva,bi2025motus,dreamzero}. These systems provide powerful behavior policies, but many remain tied to specific action representations, robot platforms, or task families; they also do not by themselves provide persistent memory, long-horizon reasoning, or runtime recovery. \modelname therefore treats these models as embodied skill backends: they execute key physical behaviors, while AgentOS schedules them, monitors their status, and recovers from failures through a shared command/status interface.

\paragraph{\calmfont{3D Spatial Memory and Scene Understanding.}}
Persistent spatial memory provides the grounding layer for deciding where embodied skills should run and what scene context they should use. Long-horizon instructions often refer to places, objects, relations, and changes that may not be visible at the current timestep. Classical 3D scene graphs organize buildings into entities and relations grounded in geometry~\citep{3d_sg_armeni_iccv19}, while open-vocabulary mapping and 3D scene-graph methods connect language queries to metric maps, dense features, object instances, or structured spatial relations~\citep{vlmaps,openscene,conceptgraphs,hovsg,ovo_mapping,iris_slam}. Recent 3D-aware embodied models further inject explicit spatial priors or view-invariant spatial representations into perception, reasoning, and action~\citep{spatialVLA,spa3r,lin2025bip3d,lin2025sem,bi2025motus}. Language-guided navigation is a representative application of scene understanding: agents must map instructions to places, landmarks, and traversable routes, as studied in instruction-following and scene-graph-based navigation settings~\citep{r2r,rxr,fsr_vln}. \modelname builds on these directions by using 3D representations as an operational memory layer: geometry supports localization and traversability, semantics supports language grounding, and scene-graph structure supports object- and room-level reasoning. Coupled with the embodied skill interface, this memory layer lets AgentOS combine VLA-style behavior models with spatial understanding during navigation, object search, manipulation, memory update, and failure recovery.

\section{Conclusion}
\label{sec:conclusion}

We presented \modelname, a unified embodied agent framework that helps narrow the embodiment gap between digital LLM agents and real-world robots. \modelname is organized through three coupled layers: Embodied AgentOS for planning, scheduling, monitoring, and re-planning; the Memory Layer for persistent spatial grounding and execution history; and the Skill Layer for structured executable robot capabilities. Quantitative experiments on semantic mapping and the \holoagentnav navigation stack show that the memory and navigation components provide competitive spatial grounding and goal-reaching performance. Real-robot demonstrations further show that the same AgentOS runtime can compose humanoid motion, object search, cross-robot coordination, and mobile manipulation through monitored skills. These results support \modelname as a practical full-stack framework for embodied-agent research and deployment, while standardized end-to-end benchmarking of heterogeneous robot skills remains an important next step.

\section{Future Work}
\label{sec:future_work}

We view \modelname as an early step toward full-stack robot-agent systems that connect language-level reasoning, embodied execution, persistent memory, and safe evaluation. Its current limitations point to three future directions.

\textbf{\calmfont{Instruction-aligned robot foundation models.}}
Current robot models still struggle with broad natural-language instructions across navigation, manipulation, whole-body motion, and user interaction. As a result, AgentOS must compose specialized backends with different interfaces and failure modes. Future robot foundation models should expose more language-aligned and composable action spaces that AgentOS can schedule, monitor, and verify.

\textbf{\calmfont{Broader embodiment support and full-stack humanoid skills.}}
The shared skill interface makes it possible to add new robot bodies, but each embodiment introduces different sensing, actuation, safety, and recovery constraints. Humanoids are especially demanding because mobility, manipulation, balance, and interaction must operate as one coupled skill stack. Future AgentOS systems should coordinate these capabilities across platforms within a single closed-loop execution context.

\textbf{\calmfont{Code generation for robot evolution.}}
A further direction is to let coding agents directly generate robot actions and execution policies from task intent, robot APIs, and environment context. Because executing generated actions on real robots can be costly and risky, we will introduce a digital-twin sandbox with EmbodiedGen~\citep{embodiedgen} for fast policy validation and evaluation before deployment. This sandbox can test whether generated actions satisfy task goals, respect embodiment constraints, and recover from controlled environment variations.

Together, these directions extend \modelname from a deployable embodied-agent framework toward a more general, embodiment-aware, and safely validated robot-agent stack.

%

\newlength{\vertsep}
\setlength{\vertsep}{.085in}
\newlength{\imsize}
\setlength{\imsize}{.365\textwidth}

\clearpage

\bibliography{paper}
\bibliographystyle{unsrt}

\clearpage

\appendix
\renewcommand{\thefigure}{A\arabic{figure}}
\renewcommand{\thetable}{A\arabic{table}}
\setcounter{figure}{0}
\setcounter{table}{0}

\clearpage

\end{document}